\newif\ifprivate\privatetrue    
\newif\ifuc\uctrue              
\privatefalse\ucfalse

\documentclass[12pt,twoside]{article}
  \newdimen\paravsp  \paravsp=1.3ex 
\usepackage[mathletters]{ucs} 
\usepackage[utf8x]{inputenc} 
\usepackage{latexsym,amsmath,amssymb,bm}
\usepackage{hyperref} 
\usepackage{graphicx} 
\usepackage{color}    

\newenvironment{keywords}{\centerline{\bf\small
Keywords}\begin{quote}\small}{\par\end{quote}\vskip 1ex}
\def\paradot#1{\vspace{\paravsp plus 0.5\paravsp minus 0.5\paravsp}\noindent{\bf\boldmath{#1.}}} 
\def\qmbox#1{{\quad\mbox{#1}\quad}} 
\def\hrefurl#1{\href{#1}{\rule{0ex}{1.7ex}\color{blue}\underline{\smash{#1}}}} 

\def\fr#1#2{{\textstyle\frac{#1}{#2}}} 
\def\frs#1#2{{^{#1}\!/\!_{#2}}} 
\def\student{\text{\sf student}}
\def\ID{\text{\sf ID}}
\def\name{\text{\sf name}}
\def\IQ{\text{\sf IQ}}
\def\grade{\text{\sf grade}}
\def\gender{\text{\sf gender}}

\def\fair{\text{\sf fair}}
\def\unfair{\text{\sf unfair}}
\def\PF{\text{\sf PF}}          
\def\CPF{\text{\sf CPF}}         
\def\WPF{\text{\sf WPF}}         
\def\UPF{\text{\sf UPF}}         

\begin{document}

\title{\vspace{-4ex}
\normalsize\sc\vskip 2mm\bf\Large\hrule height5pt \vskip 4mm
Fairness without Regret%
\footnote{Regret is meant here in a mathematical sense, and fairness could be replaced by various other criteria.}
\vskip 4mm \hrule height2pt}
\author{{\bf Marcus Hutter}\\[3mm]
\normalsize DeepMind \& ANU\\[2mm]
\normalsize \hrefurl{http://www.hutter1.net/}
}
\date{11 July 2019}
\maketitle

\begin{abstract}
A popular approach of achieving fairness in optimization problems is by constraining the solution space to ``fair'' solutions, which unfortunately typically reduces solution quality.
In practice, the ultimate goal is often an aggregate of sub-goals without a unique or best way of combining them or which is otherwise only partially known.
I turn this problem into a feature and suggest to use a parametrized objective and vary the parameters within reasonable ranges to get a {\em set} of optimal solutions, which can then be optimized using secondary criteria such as fairness without compromising the primary objective, i.e.\ without regret (societal cost).
\vspace{5ex}\def\contentsname{\centering\normalsize Contents}\setcounter{tocdepth}{1}
{\parskip=-2.7ex\tableofcontents}
\end{abstract}


\section{Introduction}\label{sec:Intro}


We consider the problem of optimizing a primary objective while also caring about a second criterion.
Before introducing our model, we need to clarify terminology:
The words ``(sub)optimal'', ``best'', ``solution quality'', ``regret'', ``(ir)relevant'', ``(in)comparable'' 
will always refer to the {\em primary} objective, henceforth called ``objective''.
On the other hand, ``fair'', ``just'', ``equitable'' will always refer to the {\em secondary} criterion.
If some of the latter aspects are relevant to the primary objective, they should be incorporated there.
While in practice there are differences in meaning of fair/just/equitable and possibly treatment, 
the difference does not matter in formalizing our basic idea, so we use the terms interchangeably.

We consider the problem of (automated) decision-making based on some (primary) objective $U:S→ℝ$.
Optimal solutions $s^*:=\arg\max_s U(s)$ sometimes appear to be unfair or unjust or not equitable.
A popular approach of achieving fairness or equality is by constraining the solution space $S$ \cite{Zafar:17,Agarwal:18}.
Sometimes (primary-objective) irrelevant attributes such as gender are used 
(e.g.\ admit the best students, but constrained by selecting at least 30\% women).
Diversity arguments have more force, if based on objective-relevant attributes, e.g.\ diversity in thinking or skills, rather than diversity in looks or genes. In the former case,  diversity is (only) an instrumental goal: 
If diversity indeed improves whatever the ultimate goal is, then it could in principle (already) be accounted for in the to-be-optimized objective, although operationally it may be easier to treat it as a constraint. 
If diversity does not positively correlate with the ultimate goal, but is desirable for other reasons, 
it can be modeled as a secondary objective or constraint.
This constraining-of-solution-space by (esp.\ irrelevant) factors is a popular approach, 
which unfortunately reduces solution quality \cite{Menon:18}.\!%
\footnote{In machine learning classification this is known as the Accuracy$↔$Fairness tradeoff.}

In practice, the ultimate goal is often a (possibly non-linear) aggregate of sub-goals, 
e.g.\ ``life goals'' include food, shelter, family, education, entertainment, health, wealth,  ... 
Few would argue there is a unique or best way of combining the different sub-goals into one objective. 

If we allow for a parametrized%
\footnote{This formulation also covers partially specified, partially observed, and imprecise objectives, but not stochastic uncertainty.} 
objective $U_{θ}$ and vary the parameters $θ$ within reasonable ranges $Θ$, we get a {\em set} of (incomparable) optimal solutions $\{s^*_θ:θ∈Θ\}$, one $s^*_θ:=\arg\max_s U_θ(s)$ for each $θ$, and can optimize within this set for secondary criteria such as fairness $F:S→ℝ$ without compromising the primary objective, i.e.\ without (societal) cost.
The optimal {\em fair} solution is $s^*_{θ^*}$, where $θ^*:=\arg\max_θ F(s^*_θ)$.
The next few pages discuss and illustrate this idea a bit more, but hardly go beyond this basic idea.

I kept this note deliberately simple and focus on the basic idea.
No probabilities, no machine learning, no fancy optimization algorithm -- yet. 
Besides an example, which is purely for illustration purpose only, 
I also don't discuss how objectives or fairness criteria could or should be chosen.
Whatever practitioners/society/ethicists deem appropriate, can be plugged in.
This work is also not about bias in the data; 
it assumes data is unbiased or has been debiased by other means \cite{Barocas:16,Calmon:17}.

The focus is on how to improve a (given) fairness criterion without compromising solution quality with respect to some (given) primary objectives, given unbiased data.

\section{Fairness as a Constraint}\label{sec:FAC}

\paradot{Optimal unconstrained solution}
Consider an optimization problem with {\it solutions space} $S$, 
and some {\it objective} quantified by an {\it utility function} 
$U:S→ℝ$. The\footnote{We assume finite $S$ and bounded $U$ to avoid distracting math subtleties.} 
\begin{equation}\label{eq:sstar}
  \text{{\it optimal solution} (by definition) is} ~~~~~~ s^*:=\arg\max_{s∈S} U(s)
\end{equation}

\begin{figure}[p!]
\begin{center}
  \includegraphics[scale=1]{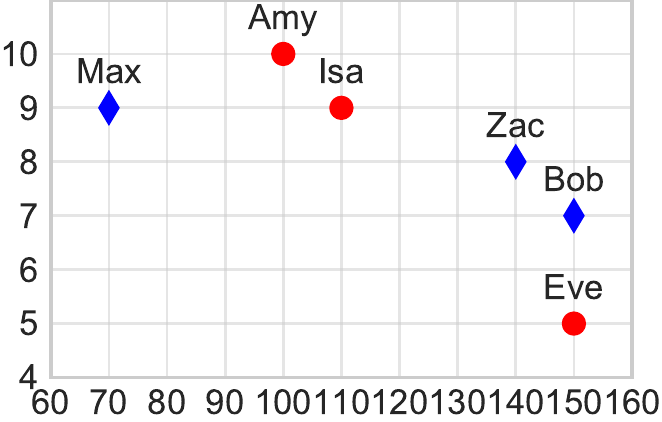}
\end{center}\vspace{-4ex}
\caption{({\bf Student example)} 6 students from $P$ with \IQ/\grade\ on horizontal/vertical axis.}\label{fig:P}\vspace*{-5ex} 
\end{figure}

\paradot{Example}
Consider a simple example of student admissions based on IQ and grade.
Assume there is a pool of 6 potential students $P≡\{\student_i: 1≤i≤6\}$ applying with information
$\student=(\ID,\name,\IQ,\grade,\gender\}$ as displayed in Table~\ref{tab:P} and Figure~\ref{fig:P}.

\begin{table}[p!]
\caption{({\bf Student data\&score)} There are 6 students in our running example $P$, together with their $θ$-weighted score $U_θ:=θ\!\cdot\!\IQ/10+(1\!-\!θ)\!\cdot\!\grade$ for various $θ$.}\label{tab:P}
\begin{center}
\begin{tabular}{c|c|c|c|c||c|c|c} 
 \ID &  \name   & \IQ & \grade & \gender & $U=U_\frs12$ & $U_{0.35}$ & $U_{0.2}$ \\ \hline
   A & \sf Amy  & 100 &   10   &    f    &      10      &{\bf 10}   &{\bf 10}    \\
   B & \sf Bob  & 150 &    7   &    m    & {\bf 11}     &    9.8    &    8.6     \\
   E & \sf Eve  & 150 &    5   &    f    &      10      &    8.5    &    7.0     \\
   I & \sf Isa  & 110 &    9   &    f    &      10      &    9.7    &{\bf 9.4}   \\
   M & \sf Max  &  70 &    9   &    m    &       8      &    8.3    &    8.6     \\
   Z & \sf Zac  & 140 &    8   &    m    & {\bf 11}     &{\bf 10.1} &    9.2     \\
\end{tabular}\vspace*{-10ex}
\end{center}
\end{table}

\begin{figure}[p!]
\begin{center}
  \includegraphics[scale=1]{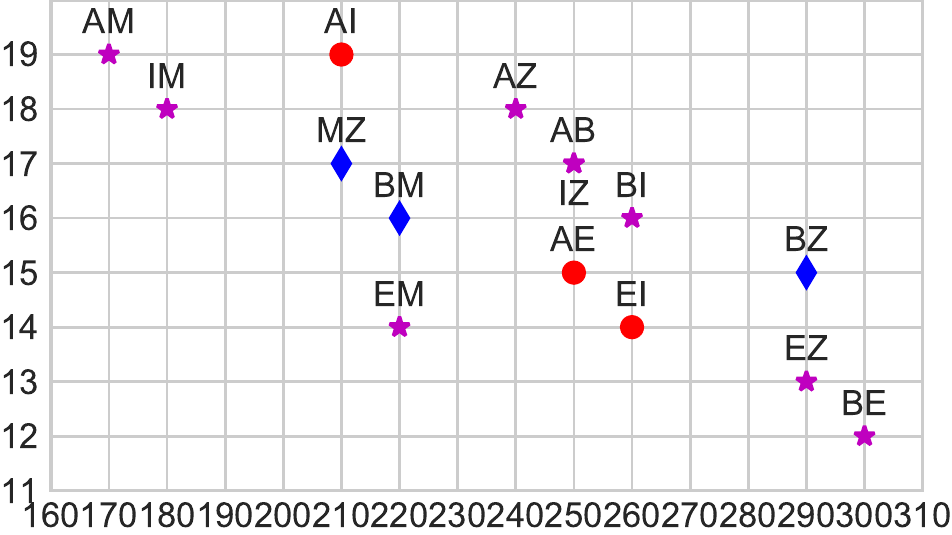}
\end{center}\vspace{-4ex}
\caption{({\bf Student pairs)} All 15 pairs of students (all potential admissions ${A⊆P:|A|=2}$) with summed \IQ/\grade\ on horizontal/vertical axis, labeled with name initials.}\label{fig:PP}
\end{figure}

Assume that high \IQ\ and \grade\ are deemed equally important for admission to University.
\IQ\ is in the range 80--150 or maybe 50--200 in general, while \grade s are in the range 5--10 or in general 0--10,
so they are not directly commensurable. Administrators typically rescale factors to make them commensurable,
so dividing \IQ\ by 10 may be adopted. We thus arrive at a performance measure 
$$
  U(\student) ~:=~ \fr12\IQ(\student)/10 + \fr12\grade(\student) ~~∈~[0;15]
$$
Assume the University can admit 2 students and $A⊆P$ is the set of potentially admitted students.
The goal then is to maximize objective 
$$
  U(A):=∑_{\student∈A}U(\student)
$$
which is the same as selecting the two students with highest $U$. The by-definition optimal selection is 
$$
  A^* ~:=~ \mathop{\arg\max}_{A⊆P:|A|=2} U(A) ~=~ \{\student∈P:U(\student)≥u\}
$$
for some suitable choice of $u$ such that the condition holds for exactly 2 students.
From the $U\!=\!U_\frs12$-column in Table~\ref{tab:P} one can see that {\sf Bob} and {\sf Zac} have the highest score,
i.e.\ $A^*=\{\sf Bob,Zac\}$.%
\!\!\footnote{To connect the notation back to (\ref{eq:sstar}), set $s:=A$ and $S=\{A⊆P:|A|=2\}$, then $s^*=A^*$. See also Figure~\ref{fig:PP}.}

\paradot{Classical fairness constraint}
In the example, the average \IQ\ and \grade\ of men is the same as for women, namely $\langle\IQ|m\rangle=120=\langle\IQ|f\rangle$ and 
$\langle\grade|m\rangle=8=\langle\grade|f\rangle$. Arguably admitting two men in this situation is unfair.%
\!\!\footnote{This is neither the place for such argument, nor what term, fair$↔$just$↔$equitable, is most appropriate.} 
Quotas have been argued to increase fairness, e.g.\ admit at least 30\% women.
Formally one restricts the solution space $S$ to fair solutions $S_\fair:=\{s∈S:F(s)=\fair\}$,
where $F:S→\{\unfair,\fair\}$ is some (exact/hard) fairness constraint, leading to the
\begin{equation}\label{eq:sstarfair}
  \text{\it optimal fair solution} ~~~~~~ s^*_\fair:=\arg\max_{s∈S_\fair} U(s)
\end{equation}

\paradot{Example}
In the student admission example, one could set $F(A):=\fair$ {\em iff} $A$ contains more than 30\% women,
in which case $A^*_\fair$ consists of {\sf Bob or Zac} and {\sf Amy or Eve or Isa}. 
All 6 solutions score the same $U(A^*_\fair)=21$ but less than $U(A^*)=22$.
In general, the constrained optimum $s^*_\fair$ is sub-optimal compared to the unconstrained optimal solution $s^*$.
Fairness comes with a cost or regret of $U(s^*) - U(s^*_\fair) > 0$ (unless $s^*_\fair=s^*$).
In the example, 10 \IQ-points or 1 in \grade\ is sacrificed.

\section{Fairness without Regret}\label{sec:FWOR}

\paradot{Uncertain objective}
The considered admission protocol involved a number of not-so-well justified steps.
For instance, \IQ\ and \grade\ were weighed equally, 
but what if overall student grade is refined to {\sf STEM} grade and {\sf HASS} grade, 
then weighing \IQ:{\sf STEM}:{\sf HASS} as 1:1:1 may be more natural, 
effectively weighing \IQ\ by $\fr13$ and \grade\ by $\fr23$.
Equally concerning is the adopted rescaling to make \IQ\ and \grade\ commensurable.
The chosen scaling was plausible but by far unique. 
Dividing \IQ\ by 20 to get \IQ\ and \grade\ into the same range 0 -- 10 seems equally justified.
While sophisticated analyzes or deliberations may narrow down the choices,
in many social real-world problems, a considerable degree of freedom or uncertainty in the objective remains.

\paradot{Core Idea: Fairness without Regret}
The main idea of this note is to actually turn this problem into a feature, 
enabling fairer decision making without regret:
If a unique objective is not achievable, consider the class of reasonable objective functions,
or at least a sub-class thereof, say $\{U_θ:θ∈Θ\}$.
Each choice $θ∈Θ$ leads to a potentially different 
\begin{equation}\label{eq:sfair1}
  \theta\text{\it -optimal solution} ~~~~ s^*_θ:=\arg\max_{s∈S} U_θ(s)
\end{equation}
Since (by assumption) no objective among $\{U_θ\}$ is more justified than another,
$U_θ$-optimal solutions $s^*_θ$ are incomparable. We hence can use some secondary criterion $F:S→ℝ$
to choose among the $U_θ$-optimal solutions $S^*_Θ:=\{s^*_θ: θ∈Θ\}$ without regret, e.g.\ the fairest solution:
\begin{equation}\label{eq:sfair2}
  s^*_{θ^*}=\arg\max_{s∈S^*_Θ}F(s) ~~~~~ (\text{with } θ^*:=\arg\max_{θ∈Θ}F(s^*_θ)~)
\end{equation}
is (the parameter corresponding to) the maximally fair solution among optimal solutions $s^*_θ$.

\paradot{Example}
In our example, we could introduce a parameter weighing \IQ\ versus \grade:
$$
  U_θ(\student) ~:=~ θ\!\cdot\!\IQ(\student)/10 ~+~ (1\!-\!θ)\!\cdot\!\grade(\student) \qmbox{with} \fr13≤θ≤\fr23
$$
We definitely want to take \IQ\ {\em and} \grade\ into account, so $θ$ should not be close to $0$ or $1$.
A range $\fr13≤θ≤\fr23$ may be deemed plausible. A smaller range seems too dogmatic while a much larger range risks to
focus too much on one attribute. The $U_θ$-optimal admissions then are 
$$
  A^*_θ ~:=~ \mathop{\arg\max}_{A⊆P:|A|=2} U_θ(A) ~=~ \{\student∈P:U_θ(\student)≥u_θ\}
$$
As a (soft/approximate) fairness criterion we could measure the male-female number mismatch
\begin{eqnarray*}
  -F(A)  & := & \Big|\#\{\student∈A:\gender(\student)=m\} \\
         &    & \!\!\!\!-~ \#\{\student∈A:\gender(\student)=f\}\Big|
\end{eqnarray*}

Table~\ref{tab:P} shows $U_θ$ for $θ=\frs12$ and $θ=0.35$ and the out-of-range $θ=0.2$.
$-F(A)$ is minimized if the number of male and female admissions is the same.
For $θ=0.35$, $A_θ=\{\sf Amy,Zac\}$ achieves this, while our original objective $U=U_\frs12$ does not.
Hence the optimal fair solution is $A^*_{θ^*}=\{\sf Amy,Zac\}$ achieved by reducing the weight of \IQ\ a bit to e.g. $0.35=θ^*∈\arg\max_θ F(A^*_θ)$. 

More generally one can show (most conveniently by inspecting Figure~\ref{fig:PP}) 
that $\frs38\!<\!θ\!<\!\frs34$ admits two men, 
$\frs14\!<\!θ^*\!<\!\frs38$ admits one male and one female, 
and $θ\!<\!\frs14$ would admit two women, 
but this has been deemed out-of-range, so no fairness criterion could achieve this, unless $Θ$ is enlarged.

Note that $U_{0.35}(A^*_{0.35})=20.1$ while $U_\frs12(A^*_\frs12)=U(A^*)=22$.
This does {\em not} imply that the fair solution is inferior to the original unconstrained solution. 
$U_θ(A)$ for different $θ$ are incomparable (even on the same $A$).  
Indeed, in general, the fair utility $U_{θ^*}(s^*_{θ^*})$ may even be higher than the original $U(s^*)$ (assuming $∃θ:U=U_θ$). 
For instance, this would happen if we added an (otherwise irrelevant) large positive constant to all \grade s,
or if we apply an (otherwise irrelevant) transformation $f_θ$ to $U_θ$, e.g.\ using $\tilde U_θ:=(1\!-\!θ)\!\cdot\!U_θ$.

\section{The Optimization Problem}\label{sec:OP}

\paradot{A naive gradient ascent algorithm}
In order to obtain an optimal fair solution $s^*_{θ^*}$ or $A^*_{θ^*}$, 
one has to solve the coupled optimization problems (\ref{eq:sfair1}) and (\ref{eq:sfair2}).
In general, this is a nasty non-convex and non-continuous double-optimization problem over discrete choices ($s∈S$ or $A⊆P$) and continuous parameters ($θ∈Θ$).
Off-the-shelf general-purpose optimization algorithms may work sometimes. 
Possibly special-purpose optimizers have to be developed for large-scale real-world problems. 

In case of a continuous solution space $S⊆ ℝ^{d'}$ 
and continuous parameter space $Θ⊆ ℝ^d$ 
and (twice) continuously differentiable $U_{\bm{θ}}(\bm{s})$ and $F(\bm{s})$, 
we could try to incrementally improve both by gradient ascent:
Assume first, we solve (\ref{eq:sfair1}) exactly, 
and want to improve fairness $F(\bm{s_θ}^*)$ by updating $\bm{θ}$ in direction of%
\footnote{All vectors are taken to be column vectors, including the gradient $\nabla$, unless transposed by $\top$.}
$$
  \nabla_{\!\bm{θ}}F(\arg\max_{\bm{s}}U_{\bm{θ}}(\bm{s})) ~≡~ \nabla_{\!\bm{θ}}F(\bm{s_θ}^*) ~=:~ {\bf G}_{\bm{θ}}(\bm{s_θ}^*)
$$
An explicit expression for $\bf G$ can be obtained by implicit differentiation \cite[Lem.1\&2]{Hutter:16vacrecog}%
\footnote{Differentiate $\nabla_{\!\bm{s}}U_{\bm{θ}}({\bm{s}})_{|{\bm{s}}={\bm{s_θ}^*}}≡0$ 
w.r.t.\ $\bm{θ}$ and solve for $\nabla_{\!\bm{θ}}{\bm{s_θ}^*}$, and plug this into 
$\nabla_{\!\bm{θ}}F({\bm{s_θ}^*})=\nabla_{\!\bm{θ}}{\bm{s_θ}^{*\top}}\!\cdot\!\nabla_{\!\bm{s}}F({\bm{s}})_{|{\bm{s}}={\bm{s_θ}^*}}$.}
$$
  {\bf G}_{\bm{θ}}(\bm{s}) ~=~ -\nabla_{\!\bm{θ}}\nabla_{\!\bm{s}}^\top U_{\bm{θ}}({\bm{s}}) \cdot
                         [\nabla_{\!\bm{s}}\nabla_{\!\bm{s}}^\top U_{\bm{θ}}({\bm{s}})]^{-1} \cdot
                          \nabla_{\!\bm{s}} F({\bm{s}})
$$
Starting with some $(\bm{s},{\bm{θ}})$, for this fixed $\bm{θ}$, 
we could now improve $\bm{s}$ by either solving maximization (\ref{eq:sfair1}) exactly for ${\bm{s}} ← \bm{s_θ}^*$ or incrementally by gradient ascent
$$
  {\bm{s}} ~←~ \Pi_S[{\bm{s}}+α\nabla_{\!\bm{s}} U_θ(\bm{s})]
$$
where $α$ is the learning rate and $\Pi_S$ a projection back into $S$.
We then update ${\bm{θ}}$ to increase fairness by
$$
\bm{θ} ~←~ \Pi_Θ[\bm{θ}+β {\bf G}_{\bm{θ}}(\bm{s})]
$$
where $β$ is a learning rate and $\Pi_Θ$ a projection back into $Θ$.
We then repeat and alternate between the two gradient steps.
This is just one (naive) suggestion how the optimization problem could be solved.
This naive algorithm may give satisfactory approximate solutions on some problems.

In our student example, $S$ is discrete, but we could try some integer relaxation.
For instance, we could represent selected students $A$ as a binary vector ${\bm{s}}∈S:=\{0,1\}^6$ with
$s_i=1$ iff $\student_i∈A$, then $U(A)≡U({\bm{s}})=∑_{i=1}^6 s_i\,U(\student_i)$. 
We could then expand $S$ to the simplex $\{{\bm{s}}∈ℝ^6:s_i≥0\,∀i ~∧~ ∑_{i=1}^6 s_i=2\}$.
Unfortunately $\nabla_{\!\bm{s}}\nabla_{\!\bm{s}}^\top U_{\bm{θ}}({\bm{s}})≡0$, 
since $U_{\bm{θ}}({\bm{s}})$ is linear in $\bm{s}$,
so the double gradient algorithm above cannot be applied.

\paradot{Multi-objective optimization and Pareto optimality \cite{Miettinen:08}}
For linearly parametrized objectives (and only for those),
there is the following relation to Pareto optimality:
In multi-objective optimization one considers $m>1$ objectives $U_1,...,U_m:S→ℝ$ over solution space $S$.
A solution $s∈S$ is called Pareto optimal {\em iff} it is not dominated by any other $s'∈S$ in the sense of
$¬∃s'∈S:[∀j:U_j(s')≥U_j(s)∧∃j:U_j(s')>U_j(s)]$. 
The Pareto front $\PF⊆ S$ is the set of all Pareto optimal $s∈S$.
All other $s∉\PF$ are clearly sub-optimal. 
Consider now the weighted sum of utilities $U_{\bm{θ}}(s):=∑_{j=1}^m θ_j U_j(s)$ with $θ_j>0\,∀j$ (strict inequality is important here). 
It is easy to see that for any $\bm{θ}>0$, $s^*_{\bm{θ}}:=\arg\max_{s∈S} U_{\bm{θ}}(s)$ is Pareto optimal.
The converse, that any $s∈\PF$ is $U_{\bm{θ}}$-optimal for some $\bm{θ}>0$ however is in general not true. It holds true if $\{(U_1(s),...,U_m(s)):s∈S\}$ is a convex set,%
\!\!\footnote{Or more generally if all points in this set lie on the boundary of its convex hull,
which may or may not be true for finite $S$.}
but for a finite data sets $S$ (e.g.\ $S=\{A⊆ P:|A|=k=2\}$) in the student example) this is {\em never} convex.%
\!\!\footnote{For instance if we admit $k=1$ student in our example, then 
$\PF=\{${\sf Amy, Bob, Isa, Zac}$\}⊊S$ are Pareto optimal, 
but ${\sf Isa}$ is {\em not} $U_{\bm{θ}}$-optimal for any $\bm{θ}>0$.}
Lacking a better term, let us call $\CPF:=\{s^*_{\bm{θ}}: \bm{θ}>0\}$ the ``convex'' Pareto front.%
\!\!\footnote{\CPF\ itself can of course not be convex, since $S$ is not a vector space, 
but even $\UPF:=\{(U_1(s),...,U_m(s)):s∈\CPF\}$ is usually not convex, 
but all points in \UPF\ lie on the boundary of the convex hull of \UPF.}
An $s∈S$ is called weakly Pareto optimal (\WPF) {\em iff} $¬∃s'∈S:[∀j:U_j(s')>U_j(s)]$. 
For fair decision making, we are only interested in reasonable mixtures $\bm{θ}∈Θ⊊(0;∞)^m$,
and the set $Θ$ may not even be an axis-aligned hypercube.
Therefore in general
$$
  \{s^*_{\bm{θ}^*}\} ~⊊~ S^*_Θ ~⊊~ \CPF ~⊊~ \PF ~⊊~ \WPF ~⊊~ S
$$
For instance, for our example one can show that all inclusions are strict (see Figure~\ref{fig:PP}):
\begin{eqnarray*}
  \text{optimal Fair solution: }      ~~s^*_{θ^*}=\!\! &               & \!\!\!\phantom{\cup} ~\{{\sf Amy,Zac}\}, \\
  \text{($Θ$)-optimal solution set: } ~~S^*_Θ    =\!\! & \{s^*_{θ^*}\} & \!\!\!\cup ~\{\{{\sf Bob,Zac}\}\}, \\
  \text{convex Pareto front: }        \CPF       =\!\! & S^*_Θ         & \!\!\!\cup ~\{\{{\sf Amy,Isa}\},\{{\sf Bob,Eve}\}\}, \\
  \text{Pareto front: }               ~~\PF      =\!\! & \CPF          & \!\!\!\cup ~\{\{{\sf Amy,Bob}\},\{{\sf Bob,Isa}\},\{{\sf Isa,Zac}\}\}, \\
  \text{weak Pareto front: }          \!\WPF     =\!\! & \PF           & \!\!\!\cup ~\{\{{\sf Eve,Zac}\}\}, \\
  \text{solution space: }             ~~~~S      =\!\! & \WPF          & \!\!\!\cup ~\{\{{\sf Amy,Eve}\},\{{\sf Bob,Eve}\},\{{\sf Max,*}\}\}
\end{eqnarray*}
Nevertheless, for linear mixtures one may use ideas from multi-objective optimization
and Pareto frontiers to narrow down the solution space to aid finding $S^*_Θ$ and ultimately $s^*_{θ^*}$.
Note though that this approach is limited to linear mixtures of objectives,
but not generally parametrized objectives $U_θ$, and even $Θ$ does not need to be a (subset of a) vector space, 
so the connection to Pareto optimality is somewhat weak.

\section{Discussion}\label{sec:Disc}

\paradot{Perfect fairness} 
I have demonstrated how to incorporate fairness as a secondary optimization criterion without compromising solution quality by exploiting that many real-life objectives cannot unambiguously be defined. 
It is important to note that if there is a binary notion of perfect fairness, 
it may not be achievable with this procedure (unlike in the simple example). 

\paradot{Controversial fairness} 
On the other hand, fairness is a notoriously contentious notion \cite{Verma:18}. 
In our example, should irrelevant birth factors be even taken into account, 
i.e.\ included in the data?
If so, then which ones and why? Gender? Skin color? Eye color? Body height? 
Should any imbalance in the pool of applicants be taken into account (not a problem in our example)?
Given there are many contradictory notions of fairness \cite[Sec.4.7]{Zhong:18},
{\em improving} (presumed) fairness is probably wiser than aiming for perfect fairness.
Our approach does the former without harming solution quality;  
even optimizing for controversial fairness notions (e.g.\ demographic parity \cite{Hardt:16,Zafar:17}) becomes unproblematic.

I also assumed that there is no bias in the data, or at least this work did not address this issue. 
While removing explicit attributes in the data regarded as irrelevant is easy, 
how to deal with implicit bias in the data is subject to ongoing research \cite{Barocas:16,Calmon:17}.
One may argue that once data is debiased, 
there is no need for secondary fairness criteria, 
but the former seems difficult to achieve or even know,
and further diversity arguments will probably always remain.

\paradot{Non-unique objectives} 
Coming up with an appropriate parametrized objective can itself be a challenge,   
but arguably this is a better/easier problem than to specify a unique objective.
Being forced to agree on a relative weighing of factors can be arduous and the result may easily be determined 
by authority or whoever shouts loudest rather than rationally by reason and deliberation.
A range of objectives seems easier to converge to. 
In the simplest case one could pool the proposed utility functions of different experts, or better, start with a large parametrized class $\{U_θ\}$, e.g.\ {\em any} (non)linear combination of attributes, then choose $Θ$ to be the convex hull of expert choices $θ_1,θ_2,θ_3,...$. 
One may lean towards a smaller range $Θ$ if the fairness criterion is controversial, or a larger range $Θ$ if fairness is deemed crucial.

\paradot{Uncertainty in data}
Consider a selection problem of $k$ items from a large(r) population $P=\{x_1,...,x_n\}$ as in the example, where $x_i∈X$ was a student record, $n=6$ and $k=2$.
Assume some attributes such as \IQ\ are missing or not precisely known,
which can be modeled as interval-valued or more generally set-valued attributes. In this case, a student record becomes a set $X_i⊆ X$, the data set becomes ${\cal P}=X_1×...×X_n$, and $P∈\cal P$ is one (arbitrary) completion or choice or imputation of attributes.%
\!\!\footnote{While in this notation $P$ strictly speaking is an $n$-tupel, we will interpret $P$ also as a set of size $n$, so that $A⊆ P$ is well-defined.} 
For each choice we can find the optimal solution and then the (supposedly) fairest choice:
\begin{equation}\label{eq:uid}
 A^*_P:=\arg\max_{|A|=k} U_P(A) \qmbox{and} 
 P^*:=\arg\max_{P\in\cal P} F(A^*_P)
\end{equation}
Despite the similarity in mathematical structure to the uncertain objective case ($Θ\widehat=\cal P$ and $θ\widehat=P$), there is a crucial difference which renders $A^*_{P^*}$ actually very biased or {\em un}fair. 
Assume that naively using mean values for uncertain attributes leads to a high proportion of male admissions. Using 
({\ref{eq:uid}) instead may indeed lead to more women being admitted, 
but inspecting $P^*$ would reveal that this has been achieved by imputing \IQ\ and \grade s at the low interval boundary for males and at the high interval end for women, which is difficult to justify as fair.
To summarize: Uncertainty in data is fundamentally different 
from uncertainty in the objective, and procedure (\ref{eq:uid}) does {\em not} lead to fair decisions. 

\section{Outlook}\label{sec:Outlook}
The basic proposed idea (possibly) can and needs to be extended in various ways: 
For instance, I have not discussed stochastic uncertainty: The data could be stochastic, 
and/or the evaluation of the objective may be stochastic. 

Many problems involve a machine learning component to solve, so there could be bias and uncertainty in the learned model. 

Possibly the most important question is how much can fairness be increased by expanding a single objective to a parametrized class, or more generally, how does $F(s^*_θ)$ depend on $Θ$. 
This will heavily depend on the problem domain, primary objective, the fairness criterion, the data, 
and how large a $Θ$ can be well-justified before it becomes an opportunity for rigging rather than fairness.
To make theoretical progress on this question, some structural assumptions on $U_θ$, $Θ$, and $F$ have to be made.

Finally, in order to obtain optimal fair solutions one has to solve a challenging non-convex and 
non-continuous double-optimization problem over discrete choices and continuous parameters.

\paradot{Acknowledgements}
I thank Iason Gabriel for valuable feedback on earlier drafts.

\def\refname{\vspace{-4ex}}
\bibliographystyle{alpha} 
\begin{small}
\newcommand{\etalchar}[1]{$^{#1}$}

\end{small}
\end{document}